# Enhanced stiffness modeling of manipulators with passive joints

Anatol Pashkevich[a,b,*], Alexandr Klimchik[a,b], Damien Chablat[b]

[a] *Ecole des Mines de Nantes, 4 rue Alfred-Kastler, Nantes 44307, France*
[b] *Institut de Recherches en Communications et en Cybernetique de Nantes, UMR CNRS 6597, 1 rue de la Noe, 44321 Nantes, France*

**Abstract**

The paper presents a methodology to enhance the stiffness analysis of serial and parallel manipulators with passive joints. It directly takes into account the loading influence on the manipulator configuration and, consequently, on its Jacobians and Hessians. The main contributions of this paper are the introduction of a non-linear stiffness model for the manipulators with passive joints, a relevant numerical technique for its linearization and computing of the Cartesian stiffness matrix which allows rank-deficiency. Within the developed technique, the manipulator elements are presented as pseudo-rigid bodies separated by multidimensional virtual springs and perfect passive joints. Simulation examples are presented that deal with parallel manipulators of the Ortholide family and demonstrate the ability of the developed methodology to describe non-linear behavior of the manipulator structure such as a sudden change of the elastic instability properties (buckling).

## 1   Introduction

Mechanical stiffness[1] of a manipulator is one of the most important indicators in performance evaluation of robotic systems [1-3]. In particular, for industrial robots where the primary target is the precise manipulation of a technological tool, the manipulator stiffness defines the positioning errors due to the external loading arising during the workpiece processing. Similarly, in industrial pick-and-place applications which are intended for simple but fast manipulations, the stiffness defines admissible velocity/acceleration while approaching the target point, in order to avoid undesirable displacements due to inertia forces [4]. Other examples include large robotic manipulators for a patient positioning in medical treatment, where elastic deformations of mechanical components under the task load (and under own link weight) is the primary source of positioning errors [5]. It is obvious that in all of these cases, the desired stiffness should be high enough to meet the requirements of the relevant application.

In contrast, for service robots, which interact directly with humans, a rather low stiffness is required to eliminate collisions causing an operator injury. This leads to special "soft" (bionic or biologically inspired) manipulator architectures that are based on utilization of numerous elastic links, joints and actuators [6,7]. Some more recent examples incorporating elastic elements, are wire-driven (cable-based) robots that are potentially interesting for medical rehabilitation and rescue operations [8]. Another active research area that needs sophisticated stiffness analysis is micromanipulation where the motions are generated due to piezoelectric actuators but conventional passive joints are replaced with elastic hinges [9-11].

---

[1] Here, we distinguish the mechanical stiffness of the manipulator (passive compliance) and the stiffness of the robotic system, which may be adjusted by regulating the virtual elasticity of actuators due to dedicated control algorithm (active compliance).

* Corresponding author. Tel. +33 251 85 83 00; fax. +33 251 85 83 49;
  E-mail address: Anatol.Pashkevich@emn.fr (A. Pashkevich).

Similar to general structural mechanics [12,13], the robot stiffness analysis evaluates the manipulator resistance to the deformations caused by an external force or torque applied to the end-effector [14]. Numerically, this property is usually defined through the *stiffness matrix* $\mathbf{K}$, which is incorporated in a linear relation between the translational/rotational displacement and static forces/torques causing this transition (assuming that all of them are small enough). The inverse of $\mathbf{K}$ is usually called the *compliance matrix* and is denoted as $\mathbf{k}$. As it follows from related works, for conservative systems that are considered in this paper, $\mathbf{K}$ is 6×6 semi-definite non-negative symmetrical matrix[2] but its structure may be non-diagonal to represent the coupling between the translation and rotation [22].

In robotics, because of some specificity, the matrix $\mathbf{K}$ is usually referred to as the "Cartesian Stiffness Matrix" and it is distinguished from the "Joint-Space Stiffness Matrix" $\mathbf{K}_\theta$ that describes the relationship between the static forces/torques and corresponding deflections in the joints, [23]. Both of these stiffness matrices can be mapped to each other using the Conservative Congruency Transformation [24,25], which is trivial if the external (or internal) loading is absent.

The existing approaches for the manipulator stiffness modeling may be roughly divided into three main groups: (i) the Finite Element Analysis (FEA), (ii) the Matrix Structural Analysis (SMA), and (iii) the Virtual Joint Method (VJM) that is often called the lumped modeling. The most accurate of them is the FEA-based technique [26-30], which allows modeling links and joints with their true dimension and shape. However, it is usually applied at the final design stage because of the high computational expenses required for the repeated remeshing of the complicated 3D structure over the whole workspace [31-33]. The SMA [34-37] also incorporates the main ideas of the FEA, but operates with rather large elements – 3D flexible beams that are presented in the manipulator structure. This leads obviously to the reduction of the computational expenses, but does not provide clear physical relations required for the parametric stiffness analysis. And finally, the VJM method is based on the extension of the traditional rigid model by adding the virtual joints (localized springs), which describe the elastic deformations of the links, joints and actuators. The VJM technique is widely used at the pre-design stage and will be further developed in this paper.

It should be stressed that conventional stiffness analysis of robotic manipulators focuses on so-called *unloaded mode*, which ignores influence of the external or internal forces applied to the end-effector or to the joints. Consequently, relevant techniques are targeted at the linearization of the "force-deflection" relation in the neighborhood of the non-loaded equilibrium, which is topologically trivial and is perfectly described by the stiffness matrix. However, since many practical applications implicitly assume external and/or internal loading, the existing techniques must be extended to the case of the *loaded mode* where the manipulator may demonstrate essentially non-linear behavior, which is not exposed in the unloaded case. In particular, as it follows from general theory of elasticity, the loading may potentially lead to multiple equilibriums, to bifurcations of the equilibriums and to static instability of certain manipulator configurations. Some aspects of multiple equilibrium problem for robotic manipulators was examined in the works of Duffy et. al. [38, 39] who applied the Catastrophe theory for the stability analysis of the planar parallel manipulators with several flexural elements under external loading. In structural mechanics, similar phenomena are well-studied for the case of "axially compressed column" that exhibits *buckling* (sudden lost of static stability and rapid change of the shape) if the loading exceeds the critical value. Nevertheless, in robotics, relevant problems did not attract much attention, mainly due to high rigidity of commercially available robots. But current trends in mechanical design of manipulators that are targeted at essential reduction of moving masses motivate relaxing this assumption and enhancement of existing stiffness analysis techniques.

Thus, this paper focuses on non-linear stiffness analysis of robotic manipulators and presents computational technique that is able to detect buckling and other non-linear phenomena in elastic behaviors of manipulators under loading. The remainder of this paper is organized as follows. Section 2 presents detailed analysis of existing results and their limitations. Section 3 defines the research problem and formulates basic assumptions. Section 4 proposes a numerical algorithm for computing of the loaded static equilibrium and its stability analysis. Section 5 focuses on the stiffness matrix evaluation taking into account the external and internal loading and presence of passive joints. Section 6 contains a set of illustrative examples that demonstrate possible nonlinear behavior of loaded serial kinematic chains and relevant parallel manipulator. Section 7 presents some discussion concerning limitations and possible extensions of the developed method. And finally, Section 8 summarizes the main results and contributions.

## 2  Related works

At the preliminary design stage, the stiffness analysis can be performed using the *virtual joint method* (VJM) that describes all types of flexibility (both distributed and lumped) by localized virtual springs located in the manipulator joints. Then, for this compliant mechanism, it is computed the Cartesian stiffness matrix that depends on current configuration (posture) of the manipulator. Mathematical background for this computation originates from the work of Salisbury [40] who derived a closed-form expression for the Cartesian stiffness matrix of a serial manipulator assuming that the mechanical elasticity is concentrated in the actuated joints. Retaining this assumption, Gosselin [41] extended this result for the case of parallel manipulators (where the links were assumed to be rigid, and the passive joints to be perfect). Further development of this approach allowed taking into account the elasticity of the links, which were presented as rigid beams supplemented by linear and torsional springs [42]. At present, there are a number of variations and simplifications of VJM, which differ in modeling assumptions and numerical techniques. In particular, it was applied to the CaPAMan, Orthoglide and H4 robots, specific variants of Stewart–Gough platform, manipulators with US/UPS legs and other kinematic architectures [43-48].

---

[2] In general, for non-conservative systems and/or some special parameterisations of end-effector location, in the loaded mode the stiffness matrix may be asymmetrical This issue is discussed in details in [15-21] and others.

In the first works, the *stiffness parameters* of the virtual springs describing the link elasticity were evaluated rather approximately, using rough presentation of the link shape by rectangular beams. Besides, it was assumed that all linear and angular deflections (compression/tension, bending, torsion) are decoupled and are presented by independent one-dimensional springs that produce a diagonal stiffness matrix of size 6×6 for each link. Afterwards, this elasticity model was enhanced by using complete 6×6 non-diagonal stiffness matrix of the cantilever beam that is known from structural mechanics. This allowed taking into account all types of the translational/rotational compliance and relevant coupling between different deflections. Further advance in this direction (applicable to the links of complicated shape) led to the *FEA-based identification technique* that involves virtual loading experiments in CAD environment and stiffness matrix estimation using dedicated numerical routines [49,50]. The latter essentially increased accuracy of the VJM-modeling while preserving its high computational efficiency. It is worth mentioning that usual high computational expenses of FEA is not a critical issue here, because it is applied only once for each link (in contrast to the straightforward FEA-modeling for the entire manipulator, which requires complete re-computing for each manipulator posture).

Another important research issue is related to *passive joints* which are widely used in parallel manipulators. In the simplest case, when number/type of geometrical constraints perfectly corresponds to number/type of passive joints, the redundant variables (passive joint coordinates) may be just eliminated from the kinematic model. This allows computing the conventional Jacobian and enables a direct application of the Salisbury formula. However, in the case of over-constrained or under-constrained manipulators, the elimination technique can not be used directly.

For *serial kinematic chains* with passive joints, the problem was solved for the general case [51]. In particular, it was proposed an algorithmic solution that extends the Salisbury formula and is able to produce the rank-deficient stiffness matrices describing "zero-resistance" of the end-effector to certain type of displacements, which do not require deflections in the virtual springs (due to presence of passive joints and/or kinematic singularity of the examined posture). Relevant technique involves inversion of dedicated square matrix of size $(n_\theta + 6) \times (n_\theta + 6)$ which is composed of the links stiffness matrices and kinematic Jacobians of both virtual springs and passive joints (here $n_\theta$ is the number of passive joints). Then, the desired Cartesian stiffness matrix is obtained by simple extraction of an appropriate $6 \times 6$ sub-matrix from the computed inverse. The main advantage of this method is its computational simplicity, since the number of the virtual springs do not influence on the size of the matrix to be inverted[3]. Besides, the method does not require manual elimination of the redundant spring corresponding to the passive joints, since this operation is inherently included in the numerical algorithm.

However, for *parallel manipulators* with passive joints, solutions were obtained for less general cases. They include "pure" parallel architectures where the base and the end platform are connected by strictly serial kinematic chains. Here, the total stiffness matrix can be presented as the sum of partial matrices corresponding to separate chains (computed using the above described technique), so the passive joints are taken into account easily. Besides, in this case the over-constraining of the mechanism does not create additional difficulties. For instance, for the *over-constrained* manipulator of Orthoglide family [52], each of the parallel chains yields the stiffness matrix of rank 4 while their aggregation gives the matrix of full rank 6. But if there exists a cross-linking between the parallel chains (like in kinematic parallelograms, for instance), this method can not be applied directly. For this case, some interesting results are presented in [53-55] where the geometrical constraints were treated in a general way but detailed computational techniques were not developed.

The majority of related works implicitly assume that the stiffness is evaluated in a quasi-static configuration with no *external or internal loading*. There are very limited number of publications that directly address the loaded working mode (or so-called "large deflections" case), where in addition to the conventional "*elastic stiffness*" in the joints it is necessary to take into account the "*geometrical stiffness*" due to the change in the manipulator configuration under the load. Although the existence of this additional stiffness component for elastic structures has been known for a long time [56], the importance of this problem for robotic manipulators has been highlighted rather recently. The most essential results in this area were obtained in [57-59] where there are presented both some theoretical issues and several case studies for serial and parallel manipulators. Several authors [60-62] addressed the problem of stiffness analysis for the manipulators with internal preloading or antagonistic actuating, but in relevant equations some of the second order kinematic derivates were neglected.

To out knowledge, the first detailed study of the second-order coefficients in stiffness analysis was done in [15,16], who took into account external loads, gravity, active and passive stiffness in actuated and unactuated joints and also considered antagonistic redundant actuation. However, the derived model requires solution of non-linear matrix equation (where the joint stiffnesses and external/internal loadings are parameters), which includes a number of first- and second-order matrix derivatives that obviously must be computed in the neighborhood of loaded equilibrium that also depends on the loadings. But this issue was not considered in details and, as result, any nonlinear effects were no detected in stiffness behavior of the examined manipulators.

---

[3] It should be noted that, in the frame of VJM approach, the Cartesian stiffness matrix may be computed for all manipulator postures, including singular ones (singular in usual sense, with respect to actuated joints). This is because the virtual joints produce the Jacobian which is always non-singular, independently of spatial location of the links. Some examples of the stiffness matrices for singular configurations can be found in [51].

**Table 1**
Summary of the related works and expressions for the Cartesian stiffness matrix

| Publications | Model & assumptions | Stiffness matrix |
|---|---|---|
| Salisbury [40] | Serial manipulator, elasticity in actuators | $\mathbf{K}_c = \mathbf{J}_\theta^{-T} \mathbf{K}_\theta \mathbf{J}_\theta^{-1}$ |
| Gosselin [41], Ciblak & Lipkin [19,23], Pigoski et al., [20] | Parallel manipulator, elasticity in actuators, non over-constrained | $\mathbf{K}_c = \mathbf{J}_\theta^{-T} \mathbf{K}_\theta \mathbf{J}_\theta^{-1}$ |
| Zhang et al. [63,64] | Serial kinematic chain without passive joints, elasticity in virtual joints | $\mathbf{K}_c = \left( \sum_i \mathbf{J}_{\theta i} \mathbf{K}_{\theta i}^{-1} \mathbf{J}_{\theta i}^T \right)^{-1}$ |
| Pashkevich et al. [49] | Serial kinematic chain with passive joints, elasticity in virtual joints | $\begin{bmatrix} \mathbf{K}_c & * \\ * & * \end{bmatrix} = \begin{bmatrix} \mathbf{J}_\theta \mathbf{K}_\theta^{-1} \mathbf{J}_\theta^T & \mathbf{J}_q \\ \mathbf{J}_q^T & \mathbf{0} \end{bmatrix}^{-1}$ |
| Zhang & Gosselin, [65] | Parallel manipulator without cross-linking between kinematic chains | $\mathbf{K}_c = \sum_i \mathbf{K}_{ci}; \quad \mathbf{K}_{ci} = \left( \sum_j \mathbf{J}_{\theta j} \mathbf{K}_{\theta j}^{-1} \mathbf{J}_{\theta j}^T \right)^{-1}$ |
| Pashkevich et al., [49] | Parallel manipulator without cross-linking between kinematic chains | $\mathbf{K}_c = \sum_i \mathbf{K}_{ci}; \quad \begin{bmatrix} \mathbf{K}_{ci} & * \\ * & * \end{bmatrix} = \begin{bmatrix} \mathbf{J}_{\theta i} \mathbf{K}_{\theta i}^{-1} \mathbf{J}_{\theta i}^T & \mathbf{J}_{qi} \\ \mathbf{J}_{qi}^T & \mathbf{0} \end{bmatrix}^{-1}$ |
| Alici & Shirinzadeh, [57] Chen & Kao [24,59], Marlet & Gosselin [66] | Serial or parallel manipulator with external loading (non over-constrained) | $\mathbf{K}_c = \mathbf{J}_\theta^{-T} \left( \mathbf{K}_\theta - \mathbf{K}_F \right) \mathbf{J}_\theta^{-1}$ |
| Quennouelle & Gosselin [53,54] | Parallel manipulator with external loading and supplementary geometric constraints (cross-linkings) | $\mathbf{K}_c = \mathbf{J}_\theta^{-T} \left( \mathbf{K}_\theta - \mathbf{K}_F + \mathbf{K}_I \right) \mathbf{J}_\theta^{-1}$ |
| Yi & Freeman [15,16] | Parallel manipulator with external loading, inertia and gravity loads, joint stiffness, actuation redundancy | $\mathbf{K}_c = \mathbf{J}_\theta^{-T} \mathbf{K}_u \mathbf{J}_\theta^{-1}$ $\mathbf{K}_u$ - solution of non-linear matrix equation that includes joint stiffnesses and external/internal loadings as parameters |

$\mathbf{K}_c$ - Cartesian stiffness at the end-effector ($6 \times 6$)

$\mathbf{K}_\theta$ - aggregated stiffness of the virtual springs ($n_\theta \times n_\theta$)

$\mathbf{J}_\theta$ - Jacobian of the virtual springs ($6 \times n_\theta$)

$\mathbf{J}_q$ - Jacobian of the passive joints ($6 \times n_q$)

$\mathbf{K}_F$ - stiffness matrix induced by external loading ($n_\theta \times n_\theta$)

$\mathbf{K}_I$ - stiffness matrix induced by constraints (cross-linking) ($n_\theta \times n_\theta$)

Therefore, in most of the related works the problem of finding the *loaded equilibrium* was omitted, so the Jacobian and Hessian were computed in a traditional way, i.e. for the unloaded configuration. The latter yielded essential computational simplification but also imposed crucial limitations, not allowing detecting the buckling and other non-linear phenomena known from general theory of elastic structures. On the other side, these issues become more and more critical in robotics applications where the geometrical buckling of the manipulator structure is more likely than the buckling in separate links.

The above mentioned publications are briefly summarised in Table 1, where all notations are adopted to those used in this paper. As it follows from their analysis, the stiffness modeling in loaded working mode for both serial and parallel manipulators with passive joints needs further improvement. In particular, it is required to develop dedicated numerical techniques that are well adopted for robotic applications and provide a designer with capability of computing the loaded equilibriums (which may be non-unique), their stability analysis, local linearization of the force-deflection relation, and also evaluation of possible non-linear phenomena (such as buckling) that can be potentially dangerous in practical applications. This motivates enhancement of our previous results [49,67,68] and extending them for the case of the external loading.

## 3    Problem of stiffness modeling

### 3.1    Basic Assumptions

The examined class of manipulators that is in the focus of this paper includes serial and parallel architectures with flexible actuated joints and flexible links that may be separated by a number of passive joints. In the case of a parallel manipulator, its architecture can be decomposed into several kinematic chains that should be studied separately. Thus, the stiffness analysis of the serial chain with passive joints (taking into account external and internal loading) is a key issue of this study. It is obvious that such kinematic chains are statically *under-constrained* and their stiffness analysis can not be performed by direct application of the standard methods.

To describe most of existing parallel architectures, it is assumed that the general serial kinematic chain under study consists of a fixed "Base", a number of flexible actuated joints "Ac", a serial chain of flexible "Links", a number of passive joints "Ps" and a moving "Platform" at the end (Fig. 1). It is also assumed that all links are separated by the joints (actuated or passive, rotational or translational) and their order is arbitrary. Besides, it is admitted that some links may be separated by actuated and passive joints simultaneously. Typical examples of the examined kinematic chains can be found in 3-PUU translational parallel kinematic machine [69], in Delta parallel robot [70] or in parallel manipulators of the Orthoglide family [52] and other manipulators[71]. It is worth mentioning that here a specific spatial arrangement of under-constrained chains yields the *over-constrained* mechanism that posses a high structural rigidity with respect to the external force.

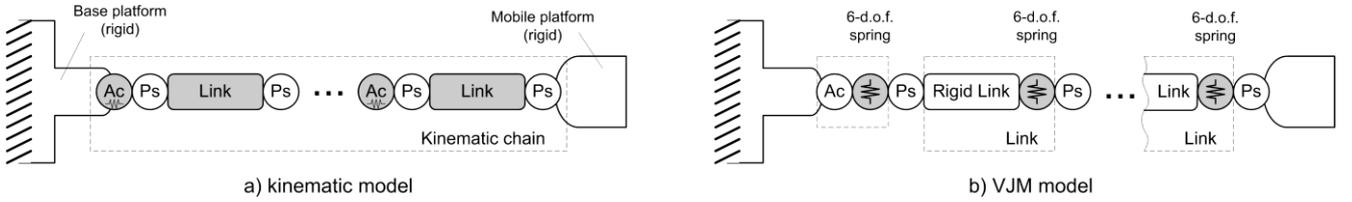

**Fig. 1.** Serial kinematic chain with passive joints and its VJM model
(Ac – actuated joint, Ps – passive joint, ⊜ - virtual spring).

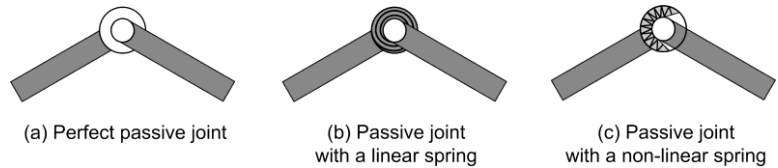

**Fig. 2.** Examples of auxiliary springs in preloaded passive joints.

To evaluate stiffness of the considered serial manipulator, let us apply a modification of the virtual joint method (VJM), which is based on the lump modeling approach [41]. According to this approach, the original rigid model should be extended by adding virtual joints (localized springs), which describe elastic deformations of the links. Besides, virtual springs are included in the actuating joints, to take into account stiffness of the control loop. Detailed description of such presentation as well as related assumptions are given in our previous publications [49, 67], where the kinematic model is presented as a product of homogeneous transformation matrices that after extraction of position and orientation of end-effector is transformed into the vector function [72]

$$\mathbf{t} = \mathbf{g}(\mathbf{q},\mathbf{\theta}) \tag{1}$$

Here the vector $\mathbf{t} = (\mathbf{p}, \mathbf{\varphi})^T$ includes the position $\mathbf{p} = (x, y, z)^T$ and orientation $\mathbf{\varphi} = (\varphi_x, \varphi_y, \varphi_z)^T$ of the end-platform (or end-effector) in the Cartesian space, the vector $\mathbf{q} = (q_1, q_2, ..., q_n)^T$ contains "perfect passive joint" coordinates (i.e., without internal preloading), the vector $\boldsymbol{\theta} = (\theta_1, \theta_2, ..., \theta_m)^T$ collects coordinates of all virtual joints and "preloaded passive joints" (with auxiliary internal springs); $n$, $m$ are the sizes of $\mathbf{q}$ and $\boldsymbol{\theta}$ respectively.

Physical nature of passive joints with internal preloading is illustrated in Fig. 2. Such joints include internal springs, so their statics is described by the following expression

$$\tau_{\theta i} = K_{\theta i} \cdot (\theta_i - \theta_{0i}) \tag{2}$$

where $\tau_{\theta i}$ is the torque/force caused by deviation of the joint coordinate $\theta_i$ from its non-loaded ("zero") value $\theta_{0i}$, and coefficient $K_{\theta i}$ defines the spring stiffness. For the purpose of generality, let us introduce similar "zero" values $\theta_{0i}$ for the virtual springs that described flexibility of the links (obviously they are equal to zero for this subset of $\boldsymbol{\theta}$). This allows to define vector $\boldsymbol{\theta}_0$ of the same size as $\boldsymbol{\theta}$ and to present the static equations corresponding to all variables (corresponding to perfect and preloaded passive joints, virtual springs of links and actuators) in general form $\boldsymbol{\tau}_\theta = \mathbf{K}_\theta \cdot (\boldsymbol{\theta} - \boldsymbol{\theta}_0)$, $\boldsymbol{\tau}_q = \mathbf{0}$. Here $\boldsymbol{\tau}_\theta$, $\boldsymbol{\tau}_q$ are the generalized torque/force in joints corresponding to the variables $\boldsymbol{\theta}$ and $\mathbf{q}$, the matrix $\mathbf{K}_\theta$ collects stiffness coefficients of all springs of the kinematic chain.

It worth mentioning that in the case without internal preloading (see for example [49, 67]) the vector $\boldsymbol{\theta}$ describes only flexibility of manipulator links/actuators that are presented by virtual springs, while vector $\mathbf{q}$ collects entire set of passive joint coordinates. In contrast, in this paper, the passive joint coordinates are divided into two subsets: (i) "perfect passive joints" included in $\mathbf{q}$, and (ii) "preloaded passive joints" included in $\boldsymbol{\theta}$ together with traditional virtual springs. Besides, if a passive joint includes a nonlinear spring (see Fig. 2), the corresponding joint variable may be included either in $\boldsymbol{\theta}$ or $\mathbf{q}$, depending on current configuration of manipulator. However, for each configuration, this assignment is strictly unique.

*3.2   Problem statement*

The designed stiffness model describes the resistance of manipulator to deformations caused by an external force or torque. It can be defined by the relation $\mathbf{F} = f(\Delta \mathbf{t})$, where $f(...)$ is a so-called "force-deflections" function that associates a deflection $\Delta \mathbf{t}$ with an external force $\mathbf{F}$ that causes the transition. It is worth mentioning that the function $f(...)$ can de determined even for the singular configurations (or redundant kinematics) while the inverse statement is not generally true. Hence, enhanced stiffness analysis must include computing of this function and detailed analysis of its singularities that may provoke various nonlinear phenomena (such as buckling). In the unloaded case, this function is usually defined through the "stiffness matrix" $\mathbf{K}$, which describes the linear relation

$$\mathbf{F} = \mathbf{K}(\mathbf{q}_0, \boldsymbol{\theta}_0) \cdot \Delta \mathbf{t} \tag{3}$$

between small six-dimensional translational/rotational displacements $\Delta \mathbf{t}$, and the external forces/torques $\mathbf{F}$ causing this transition. Here, it is assumed that $\Delta \mathbf{t}$ includes thee positional components $(\Delta x, \Delta y, \Delta z)$ describing displacement in Cartesian space and three angular components $(\Delta \varphi_x, \Delta \varphi_y, \Delta \varphi_z)$ that describe the end-platform rotation around the Cartesian axes, while the vectors $\mathbf{q}_0$, $\boldsymbol{\theta}_0$ correspond to the manipulator equilibrium for which the loadings (both internal and external) are equal to zero. However, for the loaded mode, similar linear relation is defined in the neighborhood of *another static equilibrium*, which corresponds to a different manipulator configuration $(\mathbf{q}, \boldsymbol{\theta})$, that is modified by external forces/torques $\mathbf{F}$. Respectively, in this case, the stiffness model describes the relation between the increments of the force $\delta \mathbf{F}$ and the position $\delta \mathbf{t}$

$$\delta \mathbf{F} = \mathbf{K}^F(\mathbf{q}, \boldsymbol{\theta}) \cdot \delta \mathbf{t} \tag{4}$$

where $\mathbf{q} = \mathbf{q}_0 + \Delta \mathbf{q}$ and $\boldsymbol{\theta} = \boldsymbol{\theta}_0 + \Delta \boldsymbol{\theta}$ denote the new configuration of the manipulator, and $\Delta \mathbf{q}$, $\Delta \boldsymbol{\theta}$ are the deviations in the coordinates $\mathbf{q}$, $\boldsymbol{\theta}$ respectively.

Hence, the considered problem of enhanced stiffness analysis may be divided into several sequential subtasks:

(i) *non-linear stiffness modeling* that includes computing full-scale "force-deflections relation" for the externally loaded manipulator (taking into account internal preloading) and checking stability of corresponding equilibrium configurations;

(ii) *linearization* of the relevant force-deflection relations in the neighborhood of this equilibrium and computing corresponding stiffness matrix that in general case can be singular due to presence of passive joints;

(iii) determining the *critical forces* that may cause undesired buckling phenomena or sudden change of current configuration of the loaded manipulator.

## 4 Static equilibrium for the loaded mode

Computing of the static equilibrium is a key issue for the non-linear stiffness analysis, since it defines the manipulator configuration $(\mathbf{q}, \boldsymbol{\theta})$ required for the linearization of the "load-deflection" relation. In previous works, this issue was usually ignored and the linearization was performed in the neighborhood of the unloaded configuration assuming that the external load is small enough. It is obvious that the latter essentially limits relevant results and does not allow detecting non-linear effects such as buckling. From mathematical point of view, the problem is reduced to solving of a system of non-linear static equilibrium equations that may produce unique or non-unique, stable or unstable solutions.

### *4.1 Configuration of loaded manipulator*

For computational reasons, let us consider the dual problem that deals with determining the external force $\mathbf{F}$ and the manipulator equilibrium configuration $(\mathbf{q}, \boldsymbol{\theta})$ that correspond to the end-effector location $\mathbf{t}$ taking into account internal preloading in the joints. Let us assume that the joints are given small, arbitrary virtual displacements $\delta \mathbf{q}, \delta \boldsymbol{\theta}$ in the neighborhood of $(\mathbf{q}, \boldsymbol{\theta})$. According to the principle of virtual displacements, the virtual work of the external force $\mathbf{F}$ applied to the end-effector along the corresponding displacement $\delta \mathbf{t} = \mathbf{J}_\theta \cdot \delta \boldsymbol{\theta} + \mathbf{J}_q \cdot \delta \mathbf{q}$ is equal to the sum $\left( \mathbf{F}^T \cdot \mathbf{J}_\theta \right) \cdot \delta \boldsymbol{\theta} + \left( \mathbf{F}^T \cdot \mathbf{J}_q \right) \cdot \delta \mathbf{q}$ (here $\mathbf{J}_\theta$ and $\mathbf{J}_q$ are the kinematic Jacobians with respect to the coordinates $\boldsymbol{\theta}, \mathbf{q}$). Since the passive joints do not produce the force/torque reactions, the virtual work corresponding to the generalized forces/torques $\boldsymbol{\tau}_\theta, \boldsymbol{\tau}_q$ includes only one component $-\boldsymbol{\tau}_\theta^T \cdot \delta \boldsymbol{\theta}$ (the minus sign takes into account the force-displacement directions for the virtual spring). In the static equilibrium, the total virtual work of all forces is equal to zero for any virtual displacement, therefore the equilibrium conditions may be written as $\mathbf{J}_\theta^T \cdot \mathbf{F} = \boldsymbol{\tau}_\theta$; $\mathbf{J}_q^T \cdot \mathbf{F} = \mathbf{0}$, and taking into account assumptions and notations from the previous section, the static equilibrium conditions can be presented as

$$\mathbf{J}_\theta^T \cdot \mathbf{F} = \mathbf{K}_\theta \cdot (\boldsymbol{\theta} - \boldsymbol{\theta}_0); \qquad \mathbf{J}_q^T \cdot \mathbf{F} = \mathbf{0}; \qquad \mathbf{t} = \mathbf{g}(\mathbf{q}, \boldsymbol{\theta}) \tag{5}$$

where the vector $\boldsymbol{\theta}_0$ defines internal preloading in the joints, the matrix $\mathbf{K}_\theta$ describes stiffness of all springs in the adopted VJM model, while the external loading $\mathbf{F}$ and the configuration $(\mathbf{q}, \boldsymbol{\theta})$ are treated as an unknown for given end-effector location $\mathbf{t} = \mathbf{g}(\mathbf{q}, \boldsymbol{\theta})$. Hence, the designed equilibrium configuration must satisfy the system of nonlinear equations (5).

It is evident that there is no general method for analytical solution of this system and it is required to apply numerical techniques. To derive the numerical algorithm, let us linearize the kinematic equation in the neighborhood of the current position $(\mathbf{q}_i, \boldsymbol{\theta}_i)$

$$\mathbf{t} = \mathbf{g}(\mathbf{q}_i, \boldsymbol{\theta}_i) + \mathbf{J}_q(\mathbf{q}_i, \boldsymbol{\theta}_i) \cdot (\mathbf{q}_{i+1} - \mathbf{q}_i) + \mathbf{J}_\theta(\mathbf{q}_i, \boldsymbol{\theta}_i) \cdot (\boldsymbol{\theta}_{i+1} - \boldsymbol{\theta}_i) \tag{6}$$

and rewrite the static equilibrium equations as

$$\mathbf{J}_\theta^T(\mathbf{q}_i, \boldsymbol{\theta}_i) \cdot \mathbf{F}_{i+1} = \mathbf{K}_\theta (\boldsymbol{\theta}_{i+1} - \boldsymbol{\theta}_0); \qquad \mathbf{J}_q^T(\mathbf{q}_i, \boldsymbol{\theta}_i) \cdot \mathbf{F}_{i+1} = \mathbf{0}, \tag{7}$$

which leads to a system of linear algebraic equations with respect to $(\mathbf{q}_{i+1}, \boldsymbol{\theta}_{i+1}, \mathbf{F}_{i+1})$ that includes the Jacobians $\mathbf{J}_\theta(\mathbf{q}_i, \boldsymbol{\theta}_i)$, $\mathbf{J}_q(\mathbf{q}_i, \boldsymbol{\theta}_i)$ and geometrical location function $\mathbf{g}(\mathbf{q}_i, \boldsymbol{\theta}_i)$ computed in the previous point $(\mathbf{q}_i, \boldsymbol{\theta}_i)$:

$$\begin{aligned} \mathbf{J}_q(\mathbf{q}_i, \boldsymbol{\theta}_i) \cdot \mathbf{q}_{i+1} + \mathbf{J}_\theta(\mathbf{q}_i, \boldsymbol{\theta}_i) \cdot \boldsymbol{\theta}_{i+1} &= \mathbf{t} - \mathbf{g}(\mathbf{q}_i, \boldsymbol{\theta}_i) + \mathbf{J}_q(\mathbf{q}_i, \boldsymbol{\theta}_i) \cdot \mathbf{q}_i + \mathbf{J}_\theta(\mathbf{q}_i, \boldsymbol{\theta}_i) \cdot \boldsymbol{\theta}_i; \\ \mathbf{J}_q^T(\mathbf{q}_i, \boldsymbol{\theta}_i) \cdot \mathbf{F}_{i+1} &= \mathbf{0} \\ \mathbf{J}_\theta^T(\mathbf{q}_i, \boldsymbol{\theta}_i) \cdot \mathbf{F}_{i+1} - \mathbf{K}_\theta \cdot \boldsymbol{\theta}_{i+1} &= -\mathbf{K}_\theta \cdot \boldsymbol{\theta}_0 \end{aligned} \tag{8}$$

This gives the following iterative scheme

$$\begin{bmatrix} \mathbf{F}_{i+1} \\ \mathbf{q}_{i+1} \\ \boldsymbol{\theta}_{i+1} \end{bmatrix} = \begin{bmatrix} & \mathbf{J}_q(\mathbf{q}_i, \boldsymbol{\theta}_i) & \mathbf{J}_\theta(\mathbf{q}_i, \boldsymbol{\theta}_i) \\ \mathbf{J}_q(\mathbf{q}_i, \boldsymbol{\theta}_i)^T & & \\ \mathbf{J}_\theta(\mathbf{q}_i, \boldsymbol{\theta}_i)^T & & -\mathbf{K}_\theta \end{bmatrix}^{-1} \cdot \begin{bmatrix} \mathbf{t} - \mathbf{g}(\mathbf{q}_i, \boldsymbol{\theta}_i) + \mathbf{J}_q(\mathbf{q}_i, \boldsymbol{\theta}_i) \cdot \mathbf{q}_i + \mathbf{J}_\theta(\mathbf{q}_i, \boldsymbol{\theta}_i) \cdot \boldsymbol{\theta}_i \\ \mathbf{0} \\ -\mathbf{K}_\theta \cdot \boldsymbol{\theta}_0 \end{bmatrix} \tag{9}$$

that may be reduced down to

$$\begin{bmatrix} \mathbf{F}_{i+1} \\ \mathbf{q}_{i+1} \end{bmatrix} = \begin{bmatrix} \mathbf{J}_\theta(\mathbf{q}_i,\boldsymbol{\theta}_i)\mathbf{K}_\theta^{-1}\mathbf{J}_\theta^T(\mathbf{q}_i,\boldsymbol{\theta}_i) & \mathbf{J}_q(\mathbf{q}_i,\boldsymbol{\theta}_i) \\ \mathbf{J}_q^T(\mathbf{q}_i,\boldsymbol{\theta}_i) & 0 \end{bmatrix}^{-1} \cdot \begin{bmatrix} \boldsymbol{\varepsilon}_i \\ 0 \end{bmatrix}; \qquad \boldsymbol{\theta}_{i+1} = \mathbf{K}_\theta^{-1} \cdot \mathbf{J}_\theta^T(\mathbf{q}_i,\boldsymbol{\theta}_i) \cdot \mathbf{F}_{i+1} + \boldsymbol{\theta}_0 \qquad (10)$$

where $\boldsymbol{\varepsilon}_i = \mathbf{t} - \mathbf{g}(\mathbf{q}_i,\boldsymbol{\theta}_i) + \mathbf{J}_q(\mathbf{q}_i,\boldsymbol{\theta}_i) \cdot \mathbf{q}_i + \mathbf{J}_\theta(\mathbf{q}_i,\boldsymbol{\theta}_i) \cdot \boldsymbol{\theta}_i$. The latter is more convenient computationally, since it requires inversion of a lower dimension matrix $(n+6) \times (n+6)$ instead of $(n+m+6) \times (n+m+6)$, where $n$, $m$ are the sizes of the vectors $\mathbf{q}$ and $\boldsymbol{\theta}$ respectively. For instance, for the kinematic chains of the Orthoglide manipulator (see example in section 6), expression (9) requires inversion of $34 \times 34$ matrix, while iterative scheme (10) needs inversion of $10 \times 10$ matrix only. It should be mentioned that $\mathbf{K}_\theta^{-1}$ is computed only ones, outside of the iterative loop.

Similar to other iterative schemes, convergence of this algorithm highly depends on the starting point. However, due to physical nature of the considered problem, it is possible to start iterations from the non-loaded configuration $(\mathbf{q}_0, \boldsymbol{\theta}_0)$. Besides, it is useful to modify the target point for each iteration in accordant with expression $\mathbf{t}_i = \alpha_i \cdot \mathbf{t} + (1-\alpha_i) \cdot \mathbf{t}_0$ using scalar variable $\alpha_i$ that is monotonically increasing from 0 up to 1. Another approach can be used for computing the force-deflection curve. Here, the starting point can be taken from previously computed loaded configuration corresponding to another value of deflection that is very close to the target one. As it follows from computational experiments, for typical values of deformations the proposed iterative algorithm possesses rather good convergence (3-5 iterations are usually enough if a configuration is stable). In our simulation studies, the convergence was evaluated as the weighted sum of residual norms corresponding to equations (5) and the iterations were terminated when this criterion achieved prescribe value.

However, some computational difficulties may arise in the case of buckling or in the area of multiple equilibriums, where convergence problem becomes rather critical and highly depends on the initial guess. Here, number of iterations increase significantly and computational time becomes non-negligible. To overcome these difficulties, it is proposed to modify the developed iterative scheme and repeat computations several times, with slightly modified initial points that are obtained by adding small random noise to $\mathbf{q}_0, \boldsymbol{\theta}_0$. Another option is adding small disturbances to $(\boldsymbol{\theta}_i, \mathbf{q}_i)$ at each iteration. These techniques were used in case studies presented section 6.

The proposed iterative scheme can be also slightly modified to solve the *dual problem*: computing an equilibrium configuration corresponding to given external loading $\mathbf{F}$ (instead of given $\mathbf{t}$). In this case, expressions (9), (10) are used in internal loop, while the designed algorithm is supplemented by external loop, which provides iterative searching for $\mathbf{t}$ corresponding to given $\mathbf{F}$

$$\mathbf{t}_{i+1} = \mathbf{t}_i + \mathbf{K}_i^{-1} \cdot (\mathbf{F} - \mathbf{F}_i) \qquad (11)$$

where $\mathbf{t}_i$, $\mathbf{F}_i$ and $\mathbf{K}_i$ are respectively the location, loading and stiffness matrix at the i-th iteration. It worth mentioning that the dual problem is meaningful only if the stiffness matrix $\mathbf{K}_i$ is non-singular. It is obvious that for a separate serial chain with passive joints the matrix $\mathbf{K}_i$ is always singular, while for the parallel manipulator it is usually non-singular due to specific assembling of kinematic chains. On the other hand, the original problem considered in this subsection (i.e. computing $\mathbf{F}$ corresponding to $\mathbf{t}$) is always physically meaningful and has at least one solution. More details concerning the developed algorithms are contained in web-appendix to this paper, which includes pseudo-code versions of relevant procedures [74].

*4.2 Stability of the loaded configuration*

To investigate possible non-linear effects that may be caused by external and internal loading, it is necessary to extend the notion of stability associated with the stiffness analysis. Traditionally, the stability of compliant mechanical systems (including manipulators) is defined as resistance of end-point location $\mathbf{t}$ with respect to the "disturbing" effects of external force $\mathbf{F}$ applied at this point. In such formulation, the stability is completely defined by the stiffness matrix $\mathbf{K}^F$ that describes linear relations (4) between the force and deflection deviations $\delta\mathbf{F}$, $\delta\mathbf{t}$ with respect to the values $\mathbf{F}$, $\mathbf{t}$. It is obvious that the matrix $\mathbf{K}^F$ is positive definite for the stable location $\mathbf{t}$.

However, in compliant manipulators with passive joints, the configuration $(\mathbf{q},\boldsymbol{\theta})$ corresponding to the same location $\mathbf{t}$ can be not unique. Moreover, these configurations may be both "stable" and "unstable" and correspond to different values of potential energy stored in the springs. From this point of view, it is worth to distinguish stability of end-point location $\mathbf{t}$ and stability of the corresponding configuration $(\mathbf{q},\boldsymbol{\theta})$. This issue becomes extremely important for the loaded mode, when due to kinematic redundancy caused by passive joints and excessive number of virtual springs, small disturbances in $(\mathbf{q},\boldsymbol{\theta})$ may provoke essential change of current equilibrium configuration that leads to reduction of the potential energy and transition to another equilibrium state, while keeping the same end-location. Hence, it is necessary to evaluate internal properties of the kinematic chain in the state of the loaded equilibrium that may correspond either to minimum or maximum of potential energy for fixed value of $\mathbf{t}$. Section 6 contains a number of illustrative examples that demonstrate equilibrium bifurcations while incrementing the external loading, buckling phenomena and difference between two types of stability notions mentioned above. Also, additional animated illustrations describing this issue are available in web-appendix [74].

To evaluate stability of the static equilibrium configuration $(\mathbf{q}, \boldsymbol{\theta})$, let us assume that the manipulator end-effector is fixed at the point $\mathbf{t} = (\mathbf{p}, \boldsymbol{\varphi})^T$ corresponding to the external load $\mathbf{F}$, but the joint coordinates are given small virtual displacements $\delta\mathbf{q}$, $\delta\boldsymbol{\theta}$ satisfying the geometrical constraint (1), i.e.

$$\mathbf{t} = \mathbf{g}(\mathbf{q}, \boldsymbol{\theta}); \qquad \mathbf{t} = \mathbf{g}(\mathbf{q} + \delta\mathbf{q}, \boldsymbol{\theta} + \delta\boldsymbol{\theta}) \qquad (12)$$

For these assumptions, let us compute the total virtual work in the joints that must be positive for a stable equilibrium and negative for an unstable one. To achieve the virtual configuration $(\mathbf{q} + \delta\mathbf{q}, \boldsymbol{\theta} + \delta\boldsymbol{\theta})$ and restore the equilibrium conditions, each of the joints must include virtual motors that generate the generalized forces/torques $\delta\boldsymbol{\tau}_q$, $\delta\boldsymbol{\tau}_\theta$ which satisfies the equations:

$$\begin{aligned} \mathbf{J}_\theta^T \cdot \mathbf{F} &= \mathbf{K}_\theta \cdot (\boldsymbol{\theta} - \boldsymbol{\theta}_0); & (\mathbf{J}_\theta + \delta\mathbf{J}_\theta)^T \cdot \mathbf{F} &= \mathbf{K}_\theta \cdot (\boldsymbol{\theta} - \boldsymbol{\theta}_0 + \delta\boldsymbol{\theta}) + \delta\boldsymbol{\tau}_\theta \\ \mathbf{J}_q^T \cdot \mathbf{F} &= 0; & (\mathbf{J}_q + \delta\mathbf{J}_q)^T \cdot \mathbf{F} &= \delta\boldsymbol{\tau}_q \end{aligned} \qquad (13)$$

After relevant transformations, the virtual torques may be expressed as

$$\delta\boldsymbol{\tau}_\theta = \delta(\mathbf{J}_\theta^T \cdot \mathbf{F}) - \mathbf{K}_\theta \cdot \delta\boldsymbol{\theta}; \qquad \delta\boldsymbol{\tau}_q = \delta(\mathbf{J}_q^T \cdot \mathbf{F}) \qquad (14)$$

where $\delta(...)$ denotes the differential with respect to $\delta\mathbf{q}$, $\delta\boldsymbol{\theta}$ that may be expanded via Hessians of the scalar function $\Psi = \mathbf{g}(\mathbf{q}, \boldsymbol{\theta})^T \cdot \mathbf{F}$:

$$\delta(\mathbf{J}_\theta^T \cdot \mathbf{F}) = \mathbf{H}_{\theta q}^F \cdot \delta\mathbf{q} + \mathbf{H}_{\theta\theta}^F \cdot \delta\boldsymbol{\theta}; \qquad \delta(\mathbf{J}_q^T \cdot \mathbf{F}) = \mathbf{H}_{qq}^F \cdot \delta\mathbf{q} + \mathbf{H}_{q\theta}^F \cdot \delta\boldsymbol{\theta} \qquad (15)$$

provided that

$$\mathbf{H}_{qq}^F = \partial^2\Psi/\partial\mathbf{q}^2; \qquad \mathbf{H}_{\theta\theta}^F = \partial^2\Psi/\partial\boldsymbol{\theta}^2; \qquad \mathbf{H}_{q\theta}^F = \mathbf{H}_{\theta q}^F = \partial^2\Psi/\partial\mathbf{q}\,\partial\boldsymbol{\theta} \qquad (16)$$

Further, taking into account that the virtual displacement from $(\mathbf{q}, \boldsymbol{\theta})$ to $(\mathbf{q} + \delta\mathbf{q}, \boldsymbol{\theta} + \delta\boldsymbol{\theta})$ leads to a gradual change of the virtual torques from $(\mathbf{0}, \mathbf{0})$ to $(\delta\boldsymbol{\tau}_q, \delta\boldsymbol{\tau}_\theta)$, the virtual work may be computed as a half of the corresponding scalar products

$$\delta W = -\frac{1}{2}\left(\delta\boldsymbol{\tau}_\theta^T \cdot \delta\boldsymbol{\theta} + \delta\boldsymbol{\tau}_q^T \cdot \delta\mathbf{q}\right), \qquad (17)$$

where the minus sign takes into account the adopted conventions for the positive directions of the forces and displacements. Hence, after appropriate substitutions and transformation to the matrix form, the desired stability condition may be written as

$$\delta W = -\frac{1}{2}\begin{bmatrix} \delta\boldsymbol{\theta}^T & \delta\mathbf{q}^T \end{bmatrix} \cdot \begin{bmatrix} \mathbf{H}_{\theta\theta}^F - \mathbf{K}_\theta & \mathbf{H}_{q\theta}^F \\ \mathbf{H}_{\theta q}^F & \mathbf{H}_{qq}^F \end{bmatrix} \cdot \begin{bmatrix} \delta\boldsymbol{\theta} \\ \delta\mathbf{q} \end{bmatrix} > 0 \qquad (18)$$

where $\delta\mathbf{q}$ and $\delta\boldsymbol{\theta}$ must satisfy to the geometrical constraints (12).

In order to take into account the relation between $\delta\mathbf{q}$ and $\delta\boldsymbol{\theta}$ that is imposed by (12), let us apply the first-order expansion of the function $\mathbf{g}(\boldsymbol{\theta}, \mathbf{q})$ that yields the following linear relation

$$\begin{bmatrix} \mathbf{J}_\theta & \mathbf{J}_q \end{bmatrix} \cdot \begin{bmatrix} \delta\boldsymbol{\theta} \\ \delta\mathbf{q} \end{bmatrix} = \mathbf{0} \ . \tag{19}$$

Then, applying the SVD- factorization [73] of the integrated Jacobian

$$\begin{bmatrix} \mathbf{J}_\theta & \mathbf{J}_q \end{bmatrix} = \begin{bmatrix} \mathbf{U}_\theta & \mathbf{U}_q \end{bmatrix} \cdot \begin{bmatrix} \mathbf{S}_r & \\ & \mathbf{0} \end{bmatrix} \cdot \begin{bmatrix} \mathbf{V}_\theta^T \\ \mathbf{V}_q^T \end{bmatrix} \tag{20}$$

and extracting from $\mathbf{V}_\theta$, $\mathbf{V}_q$ the sub-matrices $\mathbf{V}_\theta^o$, $\mathbf{V}_q^o$ corresponding to the zero singular values, a relevant null-space of the system (19) may be presented as

$$\delta\boldsymbol{\theta} = \mathbf{V}_\theta^o \cdot \delta\boldsymbol{\mu}; \qquad \delta\mathbf{q} = \mathbf{V}_q^o \cdot \delta\boldsymbol{\mu} \tag{21}$$

where $\delta\boldsymbol{\mu}$ is the arbitrary vector of the appropriate dimension (equal to the rank-deficiency of the integrated Jacobian). Hence, the stability condition (18) may be rewritten as an inequality

$$\delta W = -\frac{1}{2}\delta\boldsymbol{\mu}^T \cdot \begin{bmatrix} \mathbf{V}_\theta^o \\ \mathbf{V}_q^o \end{bmatrix}^T \cdot \begin{bmatrix} \mathbf{H}_{\theta\theta}^F - \mathbf{K}_\theta & \mathbf{H}_{q\theta}^F \\ \mathbf{H}_{\theta q}^F & \mathbf{H}_{qq}^F \end{bmatrix} \cdot \begin{bmatrix} \mathbf{V}_\theta^o \\ \mathbf{V}_\theta^o \end{bmatrix} \cdot \delta\boldsymbol{\mu} > 0 \tag{22}$$

that must be satisfied for all arbitrary non-zero $\delta\boldsymbol{\mu}$. In other words, the considered static equilibrium $(\mathbf{q}, \boldsymbol{\theta})$ is stable if (and only if) the matrix

$$\begin{bmatrix} \mathbf{V}_\theta^o \\ \mathbf{V}_q^o \end{bmatrix}^T \cdot \begin{bmatrix} \mathbf{H}_{\theta\theta}^F - \mathbf{K}_\theta & \mathbf{H}_{q\theta}^F \\ \mathbf{H}_{\theta q}^F & \mathbf{H}_{qq}^F \end{bmatrix} \cdot \begin{bmatrix} \mathbf{V}_\theta^o \\ \mathbf{V}_\theta^o \end{bmatrix} < 0 \tag{23}$$

is negative-definite. It is worth mentioning that the obtained result is in a good agreement with previous studies [58], where (for manipulators without passive joints) the stiffness properties were defined by the matrix $\mathbf{K}_\theta - \mathbf{H}_{\theta\theta}^F$ that evidently must be positive-definite for stable configurations. In Section 6 these results are applied for detecting bifurcations and buckling phenomena in typical serial kinematic chains and parallel manipulators.

## 5  Stiffness matrix for the loaded mode

The previous section presents technique that generally allows obtaining an exact relation between the elastic deformations and corresponding external force/torque. It is based on sequential computations of loaded equilibriums (and relevant force/torque) for various displacements of the manipulator end-point with respect to its unloaded location. However, though this relation is highly non-linear, common engineering practice operates with the stiffness matrix derived via the linearization.

To compute the desired stiffness matrix, let us consider the neighborhood of the loaded configuration and assume that the external force and the end-effector location are incremented by some small values $\delta\mathbf{F}$, $\delta\mathbf{t}$. Besides, let us assume that a new configuration also satisfies the equilibrium conditions. Hence, it is necessary to consider simultaneously two equilibriums corresponding to the manipulator state variables $(\mathbf{F}, \mathbf{q}, \boldsymbol{\theta}, \mathbf{t})$ and $(\mathbf{F} + \delta\mathbf{F}, \mathbf{q} + \delta\mathbf{q}, \boldsymbol{\theta} + \delta\boldsymbol{\theta}, \mathbf{t} + \delta\mathbf{t})$. Relevant equations of statics may be written as

$$\mathbf{J}_\theta^T \cdot \mathbf{F} = \mathbf{K}_\theta \cdot (\boldsymbol{\theta} - \boldsymbol{\theta}_0); \qquad \mathbf{J}_q^T \cdot \mathbf{F} = 0 \tag{24}$$

and

$$\begin{aligned} (\mathbf{J}_\theta + \delta\mathbf{J}_\theta)^T \cdot (\mathbf{F} + \delta\mathbf{F}) &= \mathbf{K}_\theta \cdot (\boldsymbol{\theta} - \boldsymbol{\theta}_0 + \delta\boldsymbol{\theta}); \\ (\mathbf{J}_q + \delta\mathbf{J}_q)^T \cdot (\mathbf{F} + \delta\mathbf{F}) &= 0 \end{aligned} \tag{25}$$

where $\delta\mathbf{J}_q(\mathbf{q}, \boldsymbol{\theta})$ and $\delta\mathbf{J}_\theta(\mathbf{q}, \boldsymbol{\theta})$ are the differentials of the Jacobians due to changes in $(\mathbf{q}, \boldsymbol{\theta})$. Besides, in the neighborhood of $(\mathbf{q}, \boldsymbol{\theta})$, the kinematic equation may be also presented in the linearized form:

$$\delta \mathbf{t} = \mathbf{J}_\theta(\mathbf{q}, \boldsymbol{\theta}) \cdot \delta \boldsymbol{\theta} + \mathbf{J}_q(\mathbf{q}, \boldsymbol{\theta}) \cdot \delta \mathbf{q}, \qquad (26)$$

Hence, after neglecting the high-order small terms and expanding the differentials via the Hessians of the function $\Psi = \mathbf{g}(\mathbf{q}, \boldsymbol{\theta})^T \cdot \mathbf{F}$ (similar to sub-section 4.2), equations (24), (25) may be rewritten as

$$\begin{aligned} \mathbf{J}_\theta^T(\mathbf{q}, \boldsymbol{\theta}) \cdot \delta \mathbf{F} + \mathbf{H}_{\theta q}^F(\mathbf{q}, \boldsymbol{\theta}) \cdot \delta \mathbf{q} + \mathbf{H}_{\theta\theta}^F(\mathbf{q}, \boldsymbol{\theta}) \cdot \delta \boldsymbol{\theta} = \mathbf{K}_\theta \cdot \delta \boldsymbol{\theta} \\ \mathbf{J}_q^T(\mathbf{q}, \boldsymbol{\theta}) \cdot \delta \mathbf{F} + \mathbf{H}_{qq}^F(\mathbf{q}, \boldsymbol{\theta}) \cdot \delta \mathbf{q} + \mathbf{H}_{q\theta}^F(\mathbf{q}, \boldsymbol{\theta}) \cdot \delta \boldsymbol{\theta} = \mathbf{0} \end{aligned} \qquad (27)$$

and the general relation between the increments of the state variables can be presented as

$$\begin{bmatrix} \mathbf{0} & \mathbf{J}_q & \mathbf{J}_\theta \\ \mathbf{J}_q^T & \mathbf{H}_{qq}^F & \mathbf{H}_{q\theta}^F \\ \mathbf{J}_\theta^T & \mathbf{H}_{\theta q}^F & \mathbf{H}_{\theta\theta}^F - \mathbf{K}_\theta \end{bmatrix} \cdot \begin{bmatrix} \delta \mathbf{F} \\ \delta \mathbf{q} \\ \delta \boldsymbol{\theta} \end{bmatrix} = \begin{bmatrix} \delta \mathbf{t} \\ \mathbf{0} \\ \mathbf{0} \end{bmatrix} \qquad (28)$$

The latter gives a straightforward numerical technique for computing the desired stiffness matrix: direct inversion of the matrix in the left-hand side of (28) and extracting from it the upper-left sub-matrix of size 6×6. Similarly, the matrices defining linear relations between the end-effector increment $\delta \mathbf{t}$ and the increments of the joint variables $\delta \boldsymbol{\theta}$, $\delta \mathbf{q}$ can be computed, i.e.:

$$\delta \mathbf{F} = \mathbf{K}_C \cdot \delta \mathbf{t}; \qquad \delta \boldsymbol{\theta} = \mathbf{K}_\theta \cdot \delta \mathbf{t}; \qquad \delta \mathbf{q} = \mathbf{K}_q \cdot \delta \mathbf{t} \qquad (29)$$

where

$$\begin{bmatrix} \mathbf{K}_C & * & * \\ \hline \mathbf{K}_q & * & * \\ \hline \mathbf{K}_\theta & * & * \end{bmatrix} = \begin{bmatrix} \mathbf{0} & \mathbf{J}_q & \mathbf{J}_\theta \\ \mathbf{J}_q^T & \mathbf{H}_{qq}^F & \mathbf{H}_{q\theta}^F \\ \mathbf{J}_\theta^T & \mathbf{H}_{\theta q}^F & \mathbf{H}_{\theta\theta}^F - \mathbf{K}_\theta \end{bmatrix}^{-1} \qquad (30)$$

It worth mentioning that the internal preloading (expressed by the variable $\boldsymbol{\theta}_0$) is not included in the latter expression in explicit way, but it directly influences on the variables $(\mathbf{q}, \boldsymbol{\theta})$ describing the equilibrium configuration and corresponding Jacobians and Hessians, which are elements of (30). Besides, in contrast to previous works, here it is possible to obtain supplementary matrices $\mathbf{K}_\theta, \mathbf{K}_q$ that give additional measures of the manipulator stiffness which evaluate sensitivity of the joint coordinates $(\mathbf{q}, \boldsymbol{\theta})$ with respect to the external loading.

In the case when the matrix inverse (30) is computationally hard, the variable $\delta \boldsymbol{\theta}$ can be eliminated analytically, using corresponding static equation: $\delta \boldsymbol{\theta} = \mathbf{k}_\theta^F \cdot \mathbf{J}_\theta^T \cdot \delta \mathbf{F} + \mathbf{k}_\theta^F \cdot \mathbf{H}_{\theta q}^F \cdot \delta \mathbf{q}$, where $\mathbf{k}_\theta^F = \left( \mathbf{K}_\theta - \mathbf{H}_{\theta\theta}^F \right)^{-1}$. This leads to a reduced system of matrix equations with unknowns $\delta \mathbf{F}$ and $\delta \mathbf{q}$

$$\begin{bmatrix} \mathbf{J}_\theta \cdot \mathbf{k}_\theta^F \cdot \mathbf{J}_\theta^T & \mathbf{J}_q + \mathbf{J}_\theta \cdot \mathbf{k}_\theta^F \cdot \mathbf{H}_{\theta q}^F \\ \mathbf{J}_q^T + \mathbf{H}_{q\theta}^F \cdot \mathbf{k}_\theta^F \cdot \mathbf{J}_\theta^T & \mathbf{H}_{qq}^F + \mathbf{H}_{q\theta}^F \cdot \mathbf{k}_\theta^F \cdot \mathbf{H}_{\theta q}^F \end{bmatrix} \cdot \begin{bmatrix} \delta \mathbf{F} \\ \delta \mathbf{q} \end{bmatrix} = \begin{bmatrix} \delta \mathbf{t} \\ \mathbf{0} \end{bmatrix}. \qquad (31)$$

that may be treated in the similar way, i.e. the desired stiffness matrix is also obtained by direct inversion of the matrix in the left-hand side of (31) and extracting from it the upper-left sub-matrix of size 6×6:

$$\begin{bmatrix} \mathbf{K}_C & * \\ \hline \mathbf{K}_q & * \end{bmatrix} = \begin{bmatrix} \mathbf{J}_\theta \cdot \mathbf{k}_\theta^F \cdot \mathbf{J}_\theta^T & \mathbf{J}_q + \mathbf{J}_\theta \cdot \mathbf{k}_\theta^F \cdot \mathbf{H}_{\theta q}^F \\ \mathbf{J}_q^T + \mathbf{H}_{q\theta}^F \cdot \mathbf{k}_\theta^F \cdot \mathbf{J}_\theta^T & \mathbf{H}_{qq}^F + \mathbf{H}_{q\theta}^F \cdot \mathbf{k}_\theta^F \cdot \mathbf{H}_{\theta q}^F \end{bmatrix}^{-1} \qquad (32)$$

Similar to subsection 4.1, this approach allows reducing the dimension of the inverted matrix from $(n + m + 6) \times (n + m + 6)$ to $(n + 6) \times (n + 6)$, that in the case of Orthoglide corresponds to $34 \times 34$ and $10 \times 10$ respectively.

It is worth mentioning that the structure of the latter matrix is similar to one obtained for the unloaded manipulator in [49] and differs only by Hessians that take into account the influence of the external load. It should also be noted that, because of the presence of the passive joints, the stiffness matrix of a separate serial kinematic chain is always singular, but aggregation of all the chains for a parallel manipulator produces a non-singular stiffness matrix. Some examples of such aggregation are presented in our previous papers [49, 67], where the stiffness matrices of separate kinematic chains are expressed with respect to the same end-point and the resulting stiffness matrix for the parallel manipulator is computed as a straightforward sum of these components, i.e. $\mathbf{K}_C = \sum_i \mathbf{K}_C^{(i)}$ where the index '(i)' denotes the number of kinematic chain. The later follows from the superposition principle, since the total external force corresponding to the same end-effector displacement is expressed as the sum of partial forces of the separate chains.

Further simplification can be obtained by applying the block matrix factorization technique (Frobenius equation) [75] that yields the following expression

$$\begin{aligned}\mathbf{K}_C = &\left(\mathbf{J}_\theta \cdot \mathbf{k}_\theta^F \cdot \mathbf{J}_\theta^T\right)^{-1} + \left(\mathbf{J}_\theta \cdot \mathbf{k}_\theta^F \cdot \mathbf{J}_\theta^T\right)^{-1} \cdot \left(\mathbf{J}_q + \mathbf{J}_\theta \cdot \mathbf{k}_\theta^F \cdot \mathbf{H}_{\theta q}^F\right) \times \\ &\times \left(\mathbf{H}_{qq}^F + \mathbf{H}_{q\theta}^F \cdot \mathbf{k}_\theta^F \cdot \mathbf{H}_{\theta q}^F - \left(\mathbf{J}_q^T + \mathbf{H}_{q\theta}^F \cdot \mathbf{k}_\theta^F \cdot \mathbf{J}_\theta^T\right) \cdot \left(\mathbf{J}_\theta \cdot \mathbf{k}_\theta^F \cdot \mathbf{J}_\theta^T\right)^{-1} \cdot \left(\mathbf{J}_q + \mathbf{J}_\theta \cdot \mathbf{k}_\theta^F \cdot \mathbf{H}_{\theta q}^F\right)\right)^{-1} \times \\ &\times \left(\mathbf{J}_q^T + \mathbf{H}_{q\theta}^F \cdot \mathbf{k}_\theta^F \cdot \mathbf{J}_\theta^T\right) \cdot \left(\mathbf{J}_\theta \cdot \mathbf{k}_\theta^F \cdot \mathbf{J}_\theta^T\right)^{-1}\end{aligned} \qquad (33)$$

where the first term exactly corresponds to the classical formula defining stiffness of the kinematic chain without passive joints.

Hence, the presented technique allows computing the Cartesian stiffness matrix in the presence of the external and internal loading. It generalizes our previous results both for serial kinematic chains and for parallel manipulators [49,67]. In the following Section, it will be applied to several case-studies that deal with kinematic chains employed in typical parallel manipulators and demonstrate particularities of stiffness analysis of loaded manipulator with passive joints.

## 6  Illustrative examples

To demonstrate the efficiency of the proposed technique, let us present two examples that focus on the loaded working mode and deal with (i) the stiffness analysis of a serial kinematic chain with passive joints and (ii) the stiffness analysis of a translational parallel manipulator. Within this study, it is assumed that the external wrench is acting upon the manipulator end point and the influence of the gravity is negligible. In previous works [49,67], this problem was studied for the unloaded working mode and/or for small deflections in elastic elements, so any non-linear abnormality in the manipulator behavior (buckling, etc.) was non detected. This Section includes only summary of this computational study, while more details are presented in web-appendix [74]

### 6.1  Stiffness analysis of a serial kinematic chain

First, let us consider a serial kinematic chain consisting of three similar links separated by two similar rotating actuated joints. It is assumed that the chain is a part of a parallel manipulator and it is connected to the robot base via a universal passive joint, and the end-platform connection is achieved via a spherical passive joint. For each of these configurations, let us investigate *three types of the virtual springs* corresponding to different physical assumptions concerning the stiffness properties of the actuators/links. They cover the cases, in which the main flexibility is caused by the torsion in the actuators, by the link bending, and by the combination of elementary deformations of the links.

#### 6.1.1  Geometric model

The geometry of the examined kinematic chain (Fig. 3) can be defined as $U_p R_a R_a S_p$ where R, U and S denote respectively the rotational, universal and spherical joints, and the subscripts '*p*' and '*a*' refer to the passive and active joints respectively. Using the homogeneous matrix transformations, it can be described by the equation

$$\mathbf{T} = \mathbf{R}_u(\mathbf{q}_0) \cdot \mathbf{T}_x(L) \cdot \mathbf{T}_s(\boldsymbol{\theta}_1) \cdot \mathbf{R}_z(q_{a1}) \cdot \mathbf{T}_x(L) \cdot \mathbf{T}_s(\boldsymbol{\theta}_2) \cdot \mathbf{R}_z(q_{a2}) \cdot \mathbf{T}_x(L) \cdot \mathbf{T}_s(\boldsymbol{\theta}_3) \cdot \mathbf{R}_s(\mathbf{q}_t) \qquad (34)$$

where $\mathbf{R}_z(...)$ and $\mathbf{T}_x(...)$ are the elementary rotation/translation matrices around/along the *z*- and *x*-axes, $\mathbf{R}_u(...)$ is the homogeneous rotation matrix of the universal joint (incorporating two elementary rotations), $\mathbf{R}_s(...)$ is the homogeneous rotation matrix of the spherical joint (incorporating three elementary rotations), $q_{a1}$, $q_{a2}$ are the coordinates of the actuated joints, $L$ is the length of the links, $\mathbf{q}_0$ is the coordinate vector of the universal passive joint located at the robot base, $\mathbf{q}_t$ is the coordinate vector corresponding to the passive spherical joint at the end-platform, $\mathbf{T}_s(.)$ is the homogenous matrix-function describing elastic deformations in the links and actuators (they are represented by the virtual coordinates incorporated in the vectors $\boldsymbol{\theta}_1$, $\boldsymbol{\theta}_2$, $\boldsymbol{\theta}_3$). It is obvious that this model can be easily transformed into the form $\mathbf{t} = \mathbf{g}(\mathbf{q}, \boldsymbol{\theta})$ used in the frame of the developed technique.

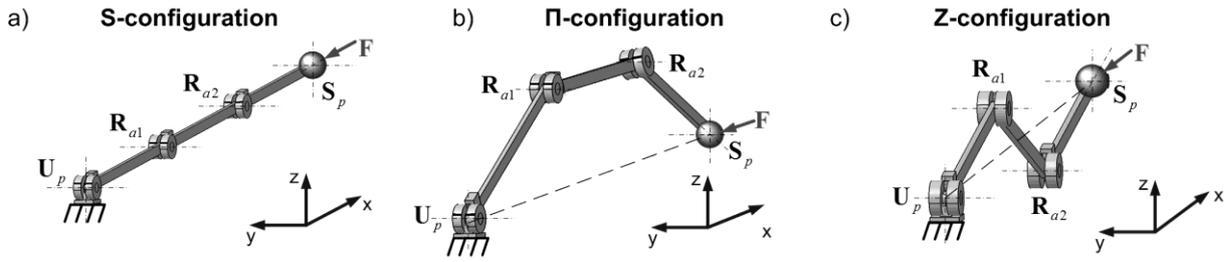

**Fig. 3.** Examined kinematical chain and its typical configurations
( $U_p$ – passive universal joint, $R_{a1}$, $R_{a2}$ – actuated rotating joints, $S_p$ – passive spherical joint)

To investigate particularities of this kinematic chain with respect to the external loading, let us also consider three typical postures that differ in values of the actuated coordinates:

- *S-configuration*: the links are located along the straight line (Fig. 3a),
  the actuated coordinates are $q_{a1} = q_{q2} = 0$
- *Π-configuration*: the chain takes a trapezoid shape (Fig. 3b),
  the actuated coordinates are $q_{a1} = q_{q2} = -30°$
- *Z-configuration*: the chain takes a zig-zag shape (Fig. 3c),
  the actuated coordinates are $q_{a1} = -q_{q2} = 30°$

For presentation convenience, let us also assume that the coordinates $\mathbf{q}_0$ of the universal passive joint are computed to ensure location of the end-effector on the Cartesian axis *x*. Besides, it is assumed that the external loading is presented by a compressing force applied at the chain end-point and it is directed along x-axis (see Fig. 3).

*6.1.2 Stiffness models*

In order to investigate possible non-linear effects in the stiffness behavior of this chain, let us consider several cases that differ in stiffness models of the links and actuated joints:

*Case of 1D-springs (Model A):* the flexible elements are localized in the actuating drives while the links are considered as strictly rigid. It allows, without loss of generality, to reduce the original $U_pR_aR_aS_p$ model down to $R_pR_aR_aR_p$ and define a single stiffness parameter $K_\theta$ (similar for both actuators) that will be used as a reference value for the further analysis. Besides, it is possible to ignore the end-effector orientation and consider a single passive joint coordinate $q$ (at the base) and two virtual joint coordinates $\theta_1$, $\theta_2$ (at actuators).

*Case of 2D springs (Model B):* the actuators do not include flexible components but the manipulator links are subject to non-negligible elastic deformations in Cartesian xy-plane (bending and compression). Correspondingly, the link flexibility is defined by a 3×3 matrix that includes elements describing deformation in x- and y- directions and rotational deformation with respect to z-axis.

*Case of 3D springs (Model C):* the actuators are strictly rigid but the link flexibility is described by a full-scale 3D model that incorporates all deflections along and around x-,y-,z-axes of the three-dimensional Cartesian space. Relevant stiffness matrix of the links has the dimension 6×6. The kinematics of this model corresponds to the general expression $U_pR_aR_aS_p$, it includes two passive joints ($\mathbf{q}$, $\mathbf{q}_s$) incorporating in total five passive coordinates and three virtual-springs with 18 virtual coordinates totally (six for each link).

*6.1.3 Stiffness analysis for model A*

In this case, the model includes minimum number of flexible elements (two 1D virtual springs in the actuated joints) and may be tackled analytically. However, in spite of its simplicity, it is potentially capable to detect the buckling phenomena at least for S-configuration, because of evident mechanical analogy to straight columns behavior under axial compression. It is also useful to evaluate other initial configurations (*Π- and Z-types*), their stability and to compare analytical solutions with numerical results provided by the developed technique.

For this model, consistent solution of geometrical and static equations (5) yields two expressions for "force-deflection" relations that correspond to stable and unstable equilibriums

$$F_s(\Delta) = \frac{K_\theta}{L} \cdot \frac{\varphi}{\sin \varphi} \ ; \qquad F_n(\Delta) = \frac{K_\theta}{L} \cdot \frac{\cos(q+\theta) + 2 \cdot \cos q}{\sin \theta} \cdot \theta \qquad (35)$$

where $F$ are the force acting along the x-axis, $\Delta$ is corresponding linear displacement, the subscripts 's' and 'n' denote manipulator configuration stability (stable or unstable), $K_\theta$ is actuator stiffness, $L$ is the length of the links, and auxiliary variables $\varphi$, $\theta$, $q$ are expressed via normalized displacement $\delta = \Delta / L$ as $\varphi = \pm \arccos(1 - \delta / 2)$, $q = \pm \arccos\left((12 - 6\delta + \delta^2)/(12 - 4\delta)\right)$, $\theta = \mp \arccos\left(1 - 3\delta/2 + \delta^2/4\right)$. It worth mentioning that other possible deformations (along y-axis) are not considered here because, as follows from separate study, in this direction the kinematic chain is kinematically singular due to passive joints and, obviously, the chain does not demonstrate any non-linear phenomena.

Corresponding force-deflections caves are presented in Fig. 4 where there are also shown the bifurcation and analytical expressions for $F$ corresponding to small values of $\Delta$ that are derived from (35) by parametric differentiation. The interpretation of this plot is similar to the axial compression of a straight column, which is a classical example in the strength of materials. It should be noted, that the developed numerical algorithm exactly produces the curve corresponding to the stable equilibrium. In this case, the external force $F \leq K_\theta / L$ can not change the manipulator shape, similar to small compressing of straight columns that can not cause lateral deflections. Consequently, for such loading the straight configuration is stable. Further, for $K_\theta / L < F \leq 3K_\theta / L$, the straight configuration may be hypothetically restored but becomes unstable, so any small disturbance will cause sudden reshaping in the direction of the stable trapezoid-type posture. And finally, for $F > 3K_\theta / L$, there may exist two types of unstable equilibriums: the trivial straight-type and a more complicated zig-zag one presented in Fig. 4.

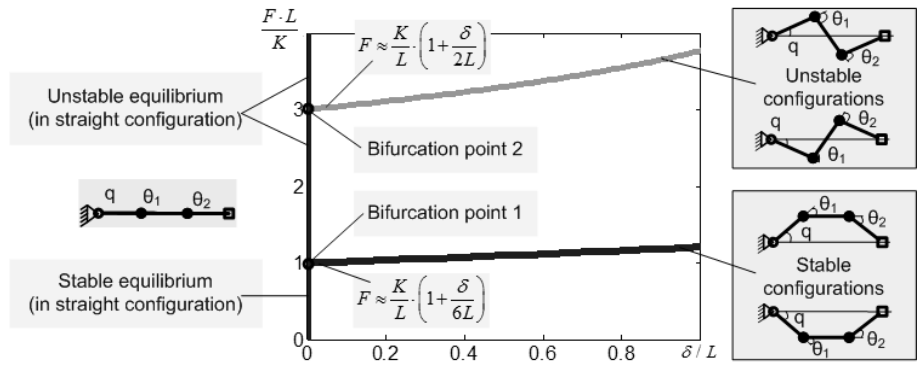

**Fig. 4.** Model A: Force-deflection relations and bifurcations for the initial S-configuration

For other initial shapes (Π-type and Z-type), the results essentially differ from the above ones (see Fig. 5). In particular, for small deflections, both Π-and Z-configuration demonstrate rather linear behavior. Moreover, in most of the cases there exist a single stable and a single unstable equilibrium, so the kinematic chain can not suddenly change its shape due to external loading. The only exception is the case of the initial Π-configuration where there are two stable and two unstable equilibriums, and here there exists a bifurcation of the stable equilibriums corresponding to the cuspidal point of the function $F(\Delta)$ where the stiffness reduces sharply. Another conclusion concerns the profile of the force-deflection plots that are highly nonlinear in all cases.

More detailed analysis shows that Π-configuration demonstrates good analogy with axially compressed imperfect column where the deflection starts from the beginning of the loading and there is no sudden buckling, but the stiffness essentially reduces while the loading increases. However, for Z-configuration that corresponds to the unloaded zig-zag shape, the stiffness behavior demonstrates the buckling that leads to sudden transformation from a symmetrical to a non-symmetrical posture as shown in Fig. 6. Here, there exist four stable equilibriums (and two unstable ones) that differ in the values of the potential energy.

Hence, in the case of model A, the developed numerical technique provides solutions that correspond to the stable loaded equilibriums and coincide with relevant analytical results. In spite of simplicity of this case study, it allows developing intuition for potentially dangerous manipulator postures with respect to buckling.

*6.1.4 Stiffness analysis for Model B*

In this case, the manipulator stiffness is caused by elasticity in the links while the actuating joints are assumed to be rigid. The elastic deflections (bending and compression) are still restricted by the Cartesian xy-plane but each link includes three virtual joints. Totally, the stiffness model has 11 variables, so it was studied numerically (using the developed technique). The stiffness parameters of the elements were evaluated assuming that the links are the rectangular beams of the length $L$ and the cross-section $a \times b$, where $a = 0.02L$ and $b = 0.05L$. For comparison purposes, corresponding stiffness matrices were scaled with respect to the bending coefficient $1/K_\theta$, to keep similarity with model A. The stiffness analysis was performed for above defined S-, Π- and Z-configurations, assuming that the external force is directed along the x-axis. Summary of the modeling results are presented in Fig. 5 and are briefly described below.

For *S-configuration*, here there is still very strong analogy with the compression of the straight column. In particular, at the beginning of the loading, the links are subject the axial compression and the stiffness is very high. Then, after the buckling, the manipulator changes its shape to become non-symmetrical and the stiffness falls down. The critical force may be also computed using the previous results, as $F_0 = K_\theta / L$. For *Π-configuration*, the stiffness properties are also qualitatively equivalent to the case of Model A but the stiffness coefficient is slightly lower. For *Z-configuration*, it was also detected the buckling that occurs if the loading approaches to the critical value $F_0 = 1.07\, K_\theta / L$.

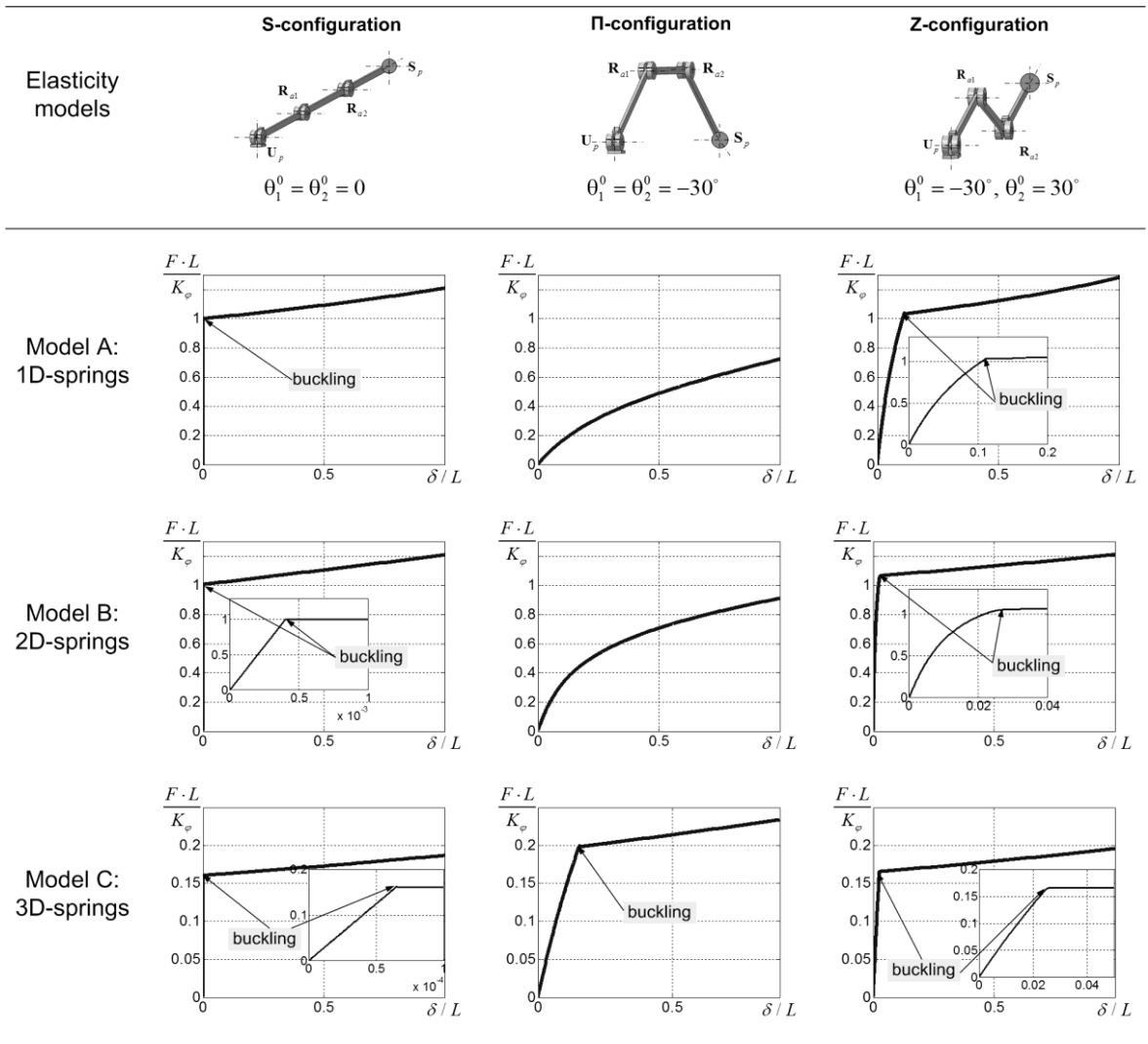

**Fig. 5.** Force-deflection relations for different elasticity models of serial chain with passive joints

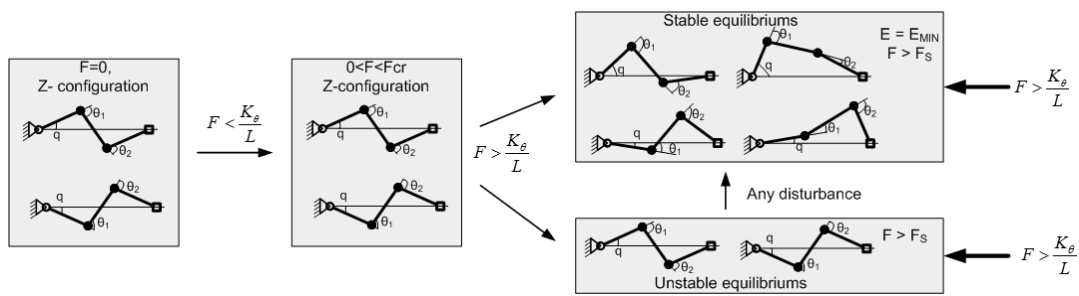

**Fig. 6.** Evolution of the Z-configuration under external loading

Hence, stiffness analysis of Model B demonstrates qualitative similarity but some quantitative difference compared to Model A. The latter is caused by different arrangement of the elastic elements in the virtual joints that corresponds to other physical assumptions.

*6.1.5 Stiffness analyses for the model C*

Finally, for model C, the link elasticity is described in 3D space and corresponding stiffness matrices have dimension 6×6 (the actuating joints are assumed perfect and rigid, similar to model B). It is also assumed that the links are rectangular beams of the length $L$ with the cross-section $a \times b$, where $a = 0.02L$, $b = 0.05L$ and the smaller value $a$ corresponds to z-direction, which was not studied above (in models A, B). To ensure comparability of all examined cases, the link stiffness matrices were parameterized with respect to the bending coefficient of the z-axis $K_\theta$. In total, the stiffness model includes 23 variables (five for passive joints and 18 for the virtual springs of three links) and it was studied numerically. The stiffness analysis was performed for the same initial configurations (S-, Π- and Z-type) and for the same direction of the external force as for the models A and B.

For *S-configuration*, the results are qualitatively similar to ones obtained for model B, but here the buckling occurs for essentially lower critical force, $0.16 K_\theta / L$, that corresponds to sudden lateral deflection in z-direction[4]. It is worth mentioning that the axial deflection corresponding to the critical force is very low (see [74] for details). In contrast, for *Π-configuration*, it was detected buckling which does not exist in models A and B. In particular, if the external force exceeds the critical value $0.20 K_\theta / L$ the stiffness suddenly reduces (it becomes 25 times lower). It corresponds to sudden deflection in z-direction that it was beyond capabilities of previous models. Another interpretation of this buckling phenomenon may be presented as sudden loss of symmetry with respect to xy-plane. For *Z-configuration*, the results remain qualitatively the same as above, but corresponding critical force is $0.17 K_\theta / L$ and it is also defined by buckling in z-direction. Therefore, model C yields essentially lower values of critical force compared to models A and B. Besides, here all examined postures demonstrated buckling in z-direction. This presents another source of potential structural instability of kinematic chains that posses the symmetry with respect to a plane.

Hence, the presented results concerning stiffness analysis of serial chains with different types of compliant elements confirm capability of the developed technique to detect (and to evaluate numerically) non-linear effects that may arise due to external loading. Besides, these examples demonstrate meaningful analogy with stiffness behavior of perfect and non-perfect columns under axial compression. It is obvious that, due to potential danger of buckling, classical procedures of manipulator design (including optimization of link cross-sections) must take into account this phenomena and the developed technique contributes to this issue.

*6.2 Stiffness analysis of translational parallel manipulator*

Let us consider now a more sophisticated example that deal with the stiffness analysis of 3-d.o.f. translational manipulator Orthoglide. This manipulator consists of three identical kinematic chains actuated by mutually orthogonal linear drives (see Fig. 7), which are arranged in a special way to ensure almost isotropic workspace kinematic properties and to restrict the end-effector motions in translation only. Such architecture ensures the velocity transmission factors close to 1.0 (similar to the conventional Cartesian-type machines) and provides high stiffness in the non-loaded mode [49,52]. However, nonlinear stiffness properties of such manipulator and potential buckling phenomena have not been studied yet.

Each kinematic chain of Orthoglide manipulator includes a foot, a kinematic parallelogram with two bars and two coupling elements, and an end-effector. The elasticity of the manipulator links was evaluated using the FEA-based methodology and special accuracy improvement tools proposed by the authors [50]. As it follows from our study, the rigidity of the parallelogram axel is high compared to the bar and the foot, thus the limb stiffness is described by equivalent 5-d.o.f. virtual spring (see[49]). For the remaining components, the compliance matrices are was taken from [50].

For the studied manipulator, the stiffness analysis was performed for vertices $Q_1…Q_4$ of the dexterous cubical workspace 200×200×200 mm$^3$ corresponding to the Orthoglide design specification with the range of the velocity transmission factors [0.5, 2.0]. Besides, similar study was done for the isotropic point $Q_0$ where the transmission factors are equal to 1.0, similar to conventional Cartesian machine [52,76]. For all of these points, there were computed non-linear curves of the force-deflection relations assuming that the displacement is directed along the bisecting line of the coordinate system (this force direction is chosen in accordance to the workpiece orientation in milling application which is currently developing for Orthoglide).

As it follows from the obtained results (Fig. 7, Table 2), for all considered points $Q_0…Q_4$, the force-deflection relations are essentially non-linear and include the buckling points where the stiffness reduces radically. Corresponding critical force varies from 2.2 kN for the point $Q_2$ to 7.8 kN for the point $Q_1$, while the critical deflections are in the range 0.5…4.9 mm. The isotropic point $Q_0$ that is usually used for the benchmarks of such manipulators possesses intermediate values of these indices (critical force 4.6 kN and critical deflection 1.4 mm).

For all cases, the translational stiffness coefficient in the pre-buckling mode is almost constant, but it essentially varies throughout the workspace. In particular, the highest value $16.6$ kN/mm is achieved at the Point $Q_1$ and the lowest one $0.4$ kN/mm corresponds to the point $Q_2$. However, after the buckling, the stiffness abruptly reduces down to $0.03…0.84$ kN/mm and further, for large deformations, it is about $0.03…0.31$ kN/mm. Hence, the traditional linear stiffness analysis (which ignores the influence of external loading) provides rather reasonable indicators for the pre-buckling mode but it does not allow evaluate the range of the loading for which the deflections may be treated as 'small' ones and the buckling phenomena do not appear yet.

---

[4] For comparison, according to the Euler formula, the local buckling of the links occurs for compressing force of 0.40 $K_\theta$/L, which is 2.5 times higher than the loading that provokes the geometric buckling of the examined mechanism.

It is worth also mentioning that for the points $Q_0...Q_4$, the physical nature of the buckling is perfectly explained by results from the previous subsection. In these cases, the buckling is caused by sudden change of the configuration of one of the serial chains (from S- or Z- to Π-configuration, for instance). It worth mentions, that parallelograms introduce some particularities that lead to additional bifurcations and salient points on the force-deflection curves. Besides, in the area of flat singularity (point $Q_2$) it is required some complimentary analysis (for more details see [74]).

Therefore, for this more sophisticated case, the developed technique also allows detecting non-linear phenomena in manipulator stiffness behavior. Moreover, most of these phenomena may be explained and predicted using presented above results for serial chains (potential dangerous configurations, corresponding critical forces and stiffness, etc.). But due to particularity of parallel manipulator geometry, it was detected another type of buckling that is caused by specific spatial location of the kinematic chains with respect to each other.

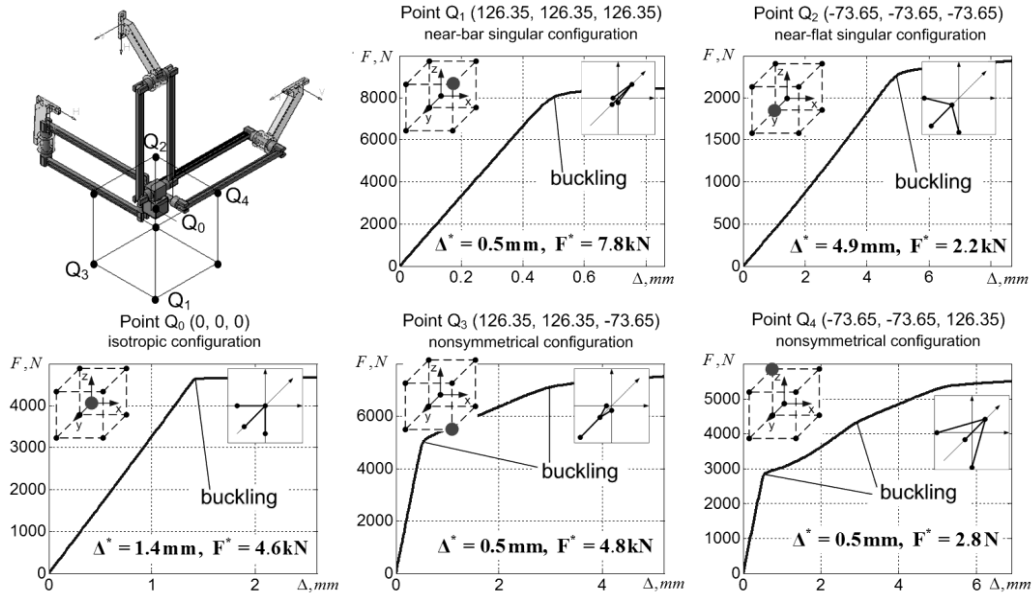

**Fig. 7.** Stiffness of the Orthoglide manipulator in critical points of the cubic workspace

**Table 2**
Summary of stiffness analysis for Orthoglide manipulator in different workpoints

| Configuration | Stiffness for unloaded mode, [kN/mm] | Critical force, [kN] | Critical deflection, [mm] | Stiffness around the buckling. [kN/mm] | | Stiffness for large deformations, [kN/mm] |
|---|---|---|---|---|---|---|
| | | | | $F < F_{cr}$ | $F > F_{cr}$ | |
| Point $Q_0$ | 3.23 | 4.65 | 1.41 | 3.36 | 0.03 | 0.03 |
| Point $Q_1$ | 16.60 | 7.85 | 0.48 | 16.40 | 0.42 | 0.31 |
| Point $Q_2$ | 0.42 | 2.24 | 4.85 | 0.48 | 0.06 | 0.03 |
| Point $Q_3$ | 9.95 | 4.75 | 0.48 | 9.60 | 0.84 | 0.11 |
| Point $Q_4$ | 5.41 | 2.78 | 0.52 | 5.09 | 0.65 | 0.06 |

## 7  Discussion

The presented results generalize previous research on the stiffness analysis of the manipulators that usually ignore the effect of the external and internal loading. Moreover, the derived expressions (30), (32) for the stiffness matrix are straightforwardly reduced to the known ones corresponding to more simple cases: (i) loaded manipulator without passive joints [24,77], and (ii) unloaded manipulator with passive joints [49]. Another advantage is that here, in contrast to previous works, the Jacobians and Hessians are computed for the joint variables corresponded to the exact loaded configuration while common approach admits that loading non-essentially influence on the Jacobian and Hessian matrices. Obviously, the proposed approach essentially improves accuracy of the stiffness model and provides it with ability of detecting various buckling phenomena.

However, there is a principal *limitation* of this method that perfectly takes into account non-linearity of the manipulator geometry but nevertheless uses linear expressions to describe the elasticity of each links (it is a fundamental assumption of all VJM-based models). As follows from our additional study, this limitation is not crucial for robotic manipulators that are design to operate in pre-buckling regime of the structural element where the link deflections are still rather small. Besides, it this research, it is assumed that the external loading is concentrated at the centre point of the end-effector while in general case the loading may be applied to the manipulator joints, or it may be distributed rather than concentrated. Another restriction is related to the manipulator architecture that is assumed to be assembled from strictly parallel serial chains (without additional geometric constraints or cross-linking between the chains). These issues are evidently left for further work.

It is also worth comparing the developed approach with other techniques that are used in neighbouring area, the *structural mechanics*. At present, there are a number of commercial finite-element software systems that are able to take into account the effects related to the geometric nonlinearity and to evaluate the critical load level and the loaded structure stability through the eigenvalue analysis of the global stiffness matrix. Usually, they implement two types of the numerical procedures: (i) computationally inexpensive "*linearized buckling analysis*" that assumes that the equilibrium point is very close to the initial configuration (sometimes it includes several iterations in the displacement space); and (ii) numerically cumbersome and straightforward "*non-linear buckling analysis*" that involves complete incremental tracing of the load-deflection path. Within both of these techniques, there are two basic versions that assume respectively that the search/tracing is performed (a) *in the displacement space* for a given fixed load vector, and (b) *in the load space* for a given fixed displacement. In commercial systems, advantages are usually given to the case (i,b) that provides good numerical stability and efficiency but affords rather moderate accuracy for the critical points.

In the frame of the above classification, the proposed technique corresponds to the case (ii,b). It allows efficiently tracing the complete load-deflection path without visible computational efforts (which is the main difficulty of the FEA-based techniques) and also admits easy incorporation of relevant iterative procedures for assessing the critical force/deflection. This numerical efficiency originates from the VJM-presentation that essentially reduces the matrix size within computations. For comparison, for the prototype of the Orthoglide manipulator created in IRCCyN (Institut de Recherche en Communications et Cybernetique de Nantes), the straightforward finite element analysis leads to the matrix inversion of size about $10^4 \times 10^4$ while the developed technique requires several inversions for matrices of size $10 \times 10$ only.

Hence, the developed technique is perfectly adapted to the stiffness analysis of the robotic manipulators. It is able to take into account influence of the external loading and allowed detecting some uncommon behavior of the robotic manipulators which was not previously reported in robotic literature. These phenomena are directly related to the robot accuracy and must be obviously taken into account during design and analysis.

# 8    Conclusions

The stiffness analysis becomes a critical issue in design optimization of robotic manipulators that currently is targeted at achieving high dynamic performances with relatively small link masses and low energy consumption in actuators. This tendency motivates a revision of existing stiffness modeling techniques that must take into account the external and internal loading. To meet this demand, the paper proposes a new systematic method for stiffness modeling of robotic manipulators with passive joints in loaded mode. It is based on a multidimensional lumped-parameter model, which presents the links as pseudo-rigid bodies with 6-d.o.f. virtual springs whose parameters are evaluated via the FEA-modeling and describe both the translational/rotational compliances and the coupling between them. The developed technique allows computing the loaded equilibrium configurations, evaluating their stability, and finding the full-scale "load-deflection" path for any given displacement of the end-effector. It is also proposed the linearization procedure for computing the Cartesian stiffness matrix, which is based on the inversion of the dedicated matrix of larger dimension, composed of the stiffness parameters of the virtual springs and the Jacobians/Hessians of the active and passive joints. These results enable designer to evaluate critical forces that may provoke non-linear behavior of the manipulators, such as sudden failure due to elastic instability (geometrical buckling) which has not been previously studied in robotic literature.

The advantages of the developed technique are illustrated by several examples that deal with parallel manipulator of the Ortholide family and serial kinematic chains employed in such robots. They demonstrate possible non-linear effects that may arise in loaded mode, including essential dependence of the stiffness on the applied force/torque and sudden change of the stiffness if the external wrench exceeds a critical value. Besides, several typical configurations of serial kinematic chains that are potentially dangerous with respect to buckling were detected.

While applied to pure parallel mechanisms with similar kinematic chains and actuators located between the base and foot, the method can be extended to other parallel manipulators to cover different actuator locations and dissimilar chain geometry. So, future work will focus on the stiffness modeling of more sophisticated architectures that include parallel manipulators with the cross-linking between the main kinematic chains and other structures that improve rigidity of the manipulating system. Another prospective research direction is the stiffness analysis of heavy manipulators, for which external loading is caused by gravity and it is distributed within the links.

# 9 Acknowledgements

The work presented in this paper was partially funded by the Region "Pays de la Loire", France and by the EU commission (project NEXT).